\documentclass{article}

     \usepackage[final]{neurips_2024}

\usepackage[utf8]{inputenc} 
\usepackage[T1]{fontenc}    
\usepackage{hyperref}       
\usepackage{url}            
\usepackage{booktabs}       
\usepackage{amsfonts}       
\usepackage{nicefrac}       
\usepackage{microtype}      
\usepackage{xcolor}         

\usepackage{microtype}
\usepackage{graphicx}
\usepackage{booktabs} 

\usepackage{hyperref}
\usepackage{bm}
\usepackage{float}
\usepackage{subfig}
\usepackage{multirow}

\usepackage{amsmath}
\usepackage{amssymb}
\usepackage{mathtools}
\usepackage{amsthm}

\usepackage{microtype}
\usepackage{graphicx}
\usepackage{booktabs} 

\usepackage{algorithm}
\usepackage{algorithmic}

\title{Take A Shortcut Back: 
           Mitigating the Gradient Vanishing for Training Spiking Neural Networks}

\author{
  Yufei Guo\thanks{Equal Contributions.}, \,
  Yuanpei Chen$^{*}$, Zecheng Hao, Weihang Peng, Zhou Jie, Yuhan Zhang, \\
  \textbf{ Xiaode Liu, Zhe Ma\thanks{Corresponding author.}} \\
  Intelligent Science \& Technology Academy of CASIC, China\\
  School of Computer Science, Peking University, China\\
  \texttt{yfguo@pku.edu.cn, rop477@163.com, haozecheng@pku.edu.cn, mazhe\_thu@163.com} \\
}

\begin{document}

\maketitle

\begin{abstract}
The Spiking Neural Network (SNN) is a biologically inspired neural network infrastructure that has recently garnered significant attention. It utilizes binary spike activations to transmit information, thereby replacing multiplications with additions and resulting in high energy efficiency. However, training an SNN directly poses a challenge due to the undefined gradient of the firing spike process. Although prior works have employed various surrogate gradient training methods that use an alternative function to replace the firing process during back-propagation, these approaches ignore an intrinsic problem: gradient vanishing. To address this issue, we propose a shortcut back-propagation method in the paper, which advocates for transmitting the gradient directly from the loss to the shallow layers. This enables us to present the gradient to the shallow layers directly, thereby significantly mitigating the gradient vanishing problem. Additionally, this method does not introduce any burden during the inference phase.
To strike a balance between final accuracy and ease of training, we also propose an evolutionary training framework and implement it by inducing a balance coefficient that dynamically changes with the training epoch, which further improves the network's performance. Extensive experiments conducted over popular datasets using several popular network structures reveal that our method consistently outperforms state-of-the-art methods.
\end{abstract}

\section{Introduction}

The Spiking Neural Network (SNN) has become a popular neural network due to its efficiency and has been widely used in various fields such as object recognition \cite{li2021free,2021TrainingXiao}, object detection \cite{kim2019spikingyolo,qu2023spiking}, and pose tracking \cite{zou2023eventbased}. The SNN operates by using binary spike signals to transmit information. When the membrane potential exceeds the threshold, the spiking neuron fires a spike represented by 1; otherwise, there is no spike represented by 0. This unique information processing paradigm is energy-efficient since it replaces multiplications of weights and activations with simple additions. Additionally, this information processing paradigm can be implemented in an efficient event-driven-based computation manner on neuromorphic hardware~\cite{2015Darwin,2015TrueNorth,2018Loihi,2019Towards,guo2023direct}, where the computational unit activates only when a spike occurs. This feature saves energy since the computational unit remains silent when there is no spike. Studies have shown that an SNN can save orders of magnitude more energy than its Artificial Neural Network (ANN) counterpart~\cite{2015TrueNorth,2018Loihi}.

Although the SNN is energy-efficient, it is challenging to train it directly because the gradient of the firing spike process is not well-defined. This means that it is impossible to use gradient-based optimization methods to train an SNN directly. To overcome this problem, researchers have proposed various surrogate gradient training (SG) methods~\citep{courbariaux2016binarized,esser2016cover,bellec2018long,2020DIET,2018Direct,2019Surrogate}. These methods use an alternative function to replace the firing process during back-propagation. For example, in~\citep{2011Error},~\citep{2017SuperSpike},~\citep{guo2022imloss}, and~\citep{2020LISNN,guo2024ternary,zhang2024enhancing}, researchers used the truncated quadratic function, the \texttt{sigmoid} function, the \texttt{tanh}-like function, and the rectangular function as surrogates, respectively. However, SG methods have an intrinsic problem: gradient vanishing. All surrogate functions are bounded, and their gradients would be close to 0 in most intervals. As a result, the gradient of the SNN would quickly decrease from output to input, causing the weights of the shallow layers of the SNN to freeze during optimization. In \autoref{sec_inf}, we will theoretically and experimentally demonstrate the gradient vanishing problem.

\begin{figure*}[t]
	\centering
	\includegraphics[width=0.9\textwidth]{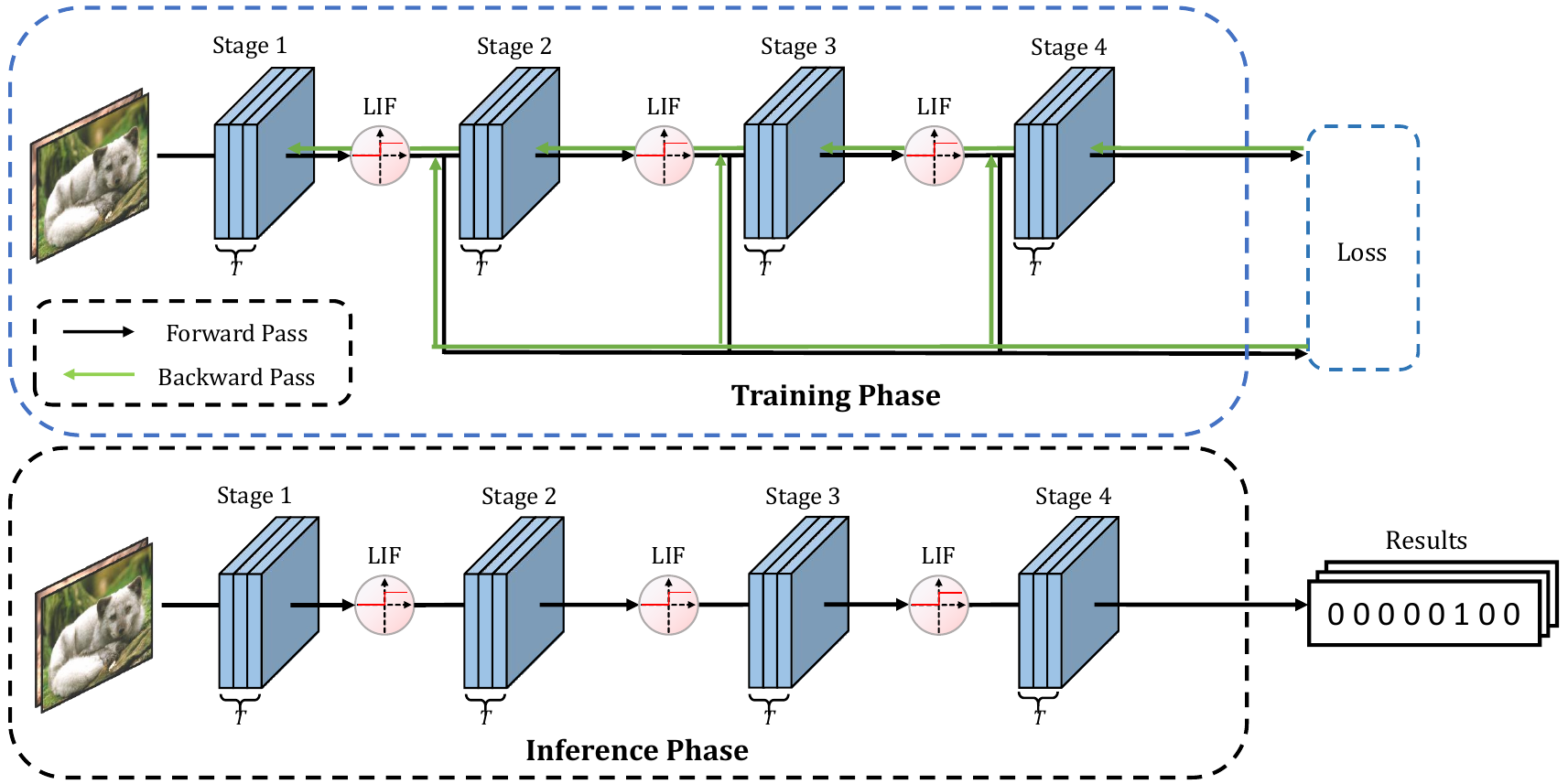} 
	\caption{The overall workflow of the proposed method. We add multiple shortcut branches from the intermediate layers to the output thus the gradient from the output could be present to the shallow layers directly.}
	\label{workflow}
\end{figure*}

To address this problem, we propose a shortcut back-propagation method in this paper, which involves transmitting the gradient from the loss to the shallow layers directly. To achieve this, we add multiple shortcut branches from intermediate layers to the output in the network. This allows information from the shallow layers to reach the output and final loss directly. Consequently, the gradient from the output can be present in the shallow layers, and their weights can be updated adequately, resulting in improved accuracy. Importantly, these shortcut branches can be removed without introducing any burden during the inference phase.
Our proposed training framework is essentially a joint optimization problem on the weighted sum of the loss functions associated with these shortcut branches. However, if we give more weight to the main branch net, the earlier layer weights may not be updated sufficiently. Conversely, if we give more attention to the side shortcut branches, the accuracy cannot reach a high level since it directly relates to the main branch outputs rather than the side branch outputs. To balance this conflict, we introduce an evolutionary training framework. During early training, we pay more attention to the side branch net, allowing sufficient weight updates of the shallow layers. Towards the end of training, we increase the weight given to the main branch net, which further improves final accuracy. We accomplish this by inducing a dynamically changing balanced coefficient that adjusts with each training epoch.
To illustrate our method's workflow, please refer to \autoref{workflow}.
Our paper provides several key contributions, which can be summarized as follows:
\begin{itemize}
\item Firstly, we have identified that the gradient vanishing problem is a significant issue for SNNs. We have supported this conclusion with theoretical justifications and in-depth experimental analysis. To mitigate this problem, we have proposed the shortcut back-propagation approach, which is a simple yet effective method. Importantly, it will not introduce any additional burden during the inference phase.

\item Secondly, we have proposed an evolutionary training framework that balances the weights of these branches with a gradual strategy. This approach ensures that earlier layer weights can be adequately updated while also improving overall accuracy. 

\item Lastly, we have evaluated the effectiveness and efficiency of our proposed methods on both static (CIFAR-10, CIFAR-100, ImageNet) and spiking (CIFAR10-DVS) datasets with widely used backbones. Our results demonstrate that the SNN trained with our proposed method is highly effective, achieving a top-1 accuracy of 77.79\% on CIFAR-100 using ResNet19 with only 2 timesteps. This represents a significant improvement of about 3.3\% compared with the current state-of-the-art SNN models even with more timesteps.
\end{itemize}

\section{Related Work}

\subsection{Learning of Spiking Neural Networks}

Unsupervised learning~\cite{Peter2015Unsupervised,2018ABiologically}, converting ANN to SNN (ANN2SNN)~\cite{2019Going,hao2023reducing,hao2023bridging},  and supervised learning~\cite{li2021differentiable,Guo_2022_CVPR} are three commonly used learning paradigms for SNNs.
In unsupervised learning, the spike-timing-dependent plasticity (STDP) approach~\cite{2020Spatial} is utilized to update the SNN model, which is considered more biologically plausible. However, due to the lack of explicit task guidance, this method is typically limited to small-scale networks and datasets.
The ANN-SNN conversion method~\cite{2020Deep,li2021free,bu2022optimal,2020TCL,2022Optimized,hao2023reducing,lan2023efficient} involves training an ANN first and then converting it into a homogeneous SNN by transferring the trained weights and replacing the ReLU neuron with a temporal spiking neuron. However, this method is not suitable for neuromorphic datasets as the ReLU neuron in the ANN cannot capture the rich temporal dynamics required for sequential information.
Supervised learning~\cite{2021Deep,2018Spatio}, on the other hand, adopts an alternative function during back-propagation to replace the firing process, enabling direct training of the SNN as an ANN. This approach leverages the success of gradient-based optimization and can achieve good performance with only a few time steps, even on large-scale datasets. Moreover, supervised learning can handle temporal data effectively, making it an increasingly popular choice in SNN research. Our work also falls within this domain.

\subsection{Relieving Training Difficulties for Supervised Learning of SNNs}

As mentioned earlier, the surrogate gradient (SG) approach is commonly employed to address the non-differentiability of SNNs. Various SG functions have been utilized, including the truncated quadratic function~\citep{2011Error}, the \texttt{sigmoid} function~\citep{2017SuperSpike}, the \texttt{tanh}-like function~\citep{guo2022imloss}, and the rectangular function~\citep{2020LISNN}.
While the SG method is generally effective, it can also introduce certain issues. Firstly, there is a gradient mismatch between the true gradient and the surrogate gradient, resulting in slow convergence and reduced accuracy. To tackle this problem, IM-Loss~\citep{guo2022imloss} proposed a dynamic manual SG method that adapts with each epoch, ensuring sufficient weight updates and accurate gradients simultaneously. In contrast to this manual design, the Differentiable Spike method~\citep{li2021differentiable} and the differentiable SG search method determine the optimal gradient estimation using finite difference and NAS techniques, respectively.
Secondly, due to the firing function being bounded, all these SG functions are also bounded. As a result, the gradient approaches or reaches close to zero in most intervals, exacerbating the vanishing gradient problem in SNNs. To mitigate this issue, SEW-ResNet~\citep{2021Deep} suggested using the ResNet with activation before addition form, while MS-ResNet~\citep{2021Advancing} advocated for the ResNet with pre-activation form. Additionally, normalization techniques have been employed to address the vanishing/explosion gradient problems. For example, Threshold-dependent batch normalization (tdBN)~\citep{2020Going} normalized the data along both the channel and temporal dimensions.
Other techniques such as Temporal Batch Normalization Through Time (BNTT)~\citep{2020Revisiting}, postsynaptic potential normalization (PSP-BN)~\citep{Rethinking2022}, and temporal effective batch normalization (TEBN)~\citep{duan2022temporal} recognized the significant variation in spike distributions across different timesteps, and thus regulated spike flows by applying separate timestep batch normalization. MPBN~\citep{Guo_2023_ICCV} introduced an additional batch normalization step after the membrane potential updating function to reestablish data flow. Similarly, regularization loss has been utilized to alleviate gradient explosion/vanishing problems. In RecDis-SNN~\citep{Guo_2022_CVPR}, a membrane potential regularization loss was proposed to control spike flow within an appropriate range.
In Spiking PointNet~\citep{ren2023spiking}, a trained-less but learning-more paradigm was proposed. This method can be seen as using a small network in the training to mitigate the training difficulty problem.

However, all these methods still need to present the gradient from the output layer to the first layer step by step, thus the gradient vanishing problem cannot be solved completely. In this paper, we propose a shortcut back-propagation method. Different from the above methods, we present the gradient from the output layer to these shallow layers directly, thus the gradient vanishing problem can be solved totally.

\section{Preliminary}


The spiking neuron serves as the fundamental and specialized computing unit in SNNs, drawing inspiration from the human brain. In our paper, we employ the widely used spiking Leaky-Integrate-and-Fire (LIF) neuron model. This model accurately captures the behavior of biological neurons by considering the interaction between the membrane potential and input current.
To show the spiking neuron in detail, we introduce the notation first.
Throughout the paper, we denote the vectors in bold italic letters. 
For instance, we use the $\bm{x}$ and $\bm{o}$ to represent the input and target output variables. 
We denote the matrices or tensors by bold capital letters (e.g., $\mathbf{W}$ is for weights). 
We denote the constants by small upright or downright letters. For example, $\bm{u}_i^{(t)}$ means the $i$-th membrane potential at time step $t$. 
Then, the LIF neuron can be described as follows: 
\begin{equation}
    \bm{u}^{(t+1), \text{pre}} = \tau\bm{u}^{(t)} + \bm{c}^{(t+1)}, \\
    \text{where } \bm{c}^{(t+1)} = \mathbf{W} \bm{x}^{(t+1)}, \label{eq_lif}
\end{equation}
where $\tau$ is a constant within $(0, 1)$, which controls the leakage of membrane potential. When $\tau$ is 1, the neuron will degenerate to the Integrate-and-Fire (IF) neuron model. In the paper, we set  $\tau$ as 0.5. $\bm{u}^{(t+1), \text{pre}}$ is the pre-synaptic input at time step $t+1$, which is charged by the input current $\bm{c}^{(t+1)}$ . Note that we omit the layer index for simplicity.
The input current is computed by the dot-product between the weights, $\mathbf{W}$ of the current layer and the spike output, $\bm{x}^{(t+1)}$ from the previous layer.
Once the membrane potential, $\bm{u}^{(t+1), \text{pre}}$ exceeds the firing threshold $V_{\rm th}$, a spike will be fired from the LIF neuron, given by
\begin{equation}
\begin{split}
    \bm{o}^{(t+1)} = 
    \begin{cases}
        1 & \text{if } \bm{u}^{(t+1), \text{pre}} > V_{\rm th} \\
        0 & \text{otherwise}
    \end{cases}, 
    \\
    \bm{u}^{(t+1)} = \bm{u}^{(t+1), \text{pre}}\cdot(1 - \bm{o}^{(t+1)}). \label{eq_fire}
\end{split}
\end{equation}
After firing, the spike output $\bm{o}^{(t+1)}$ will propagate to the next layer and become the input $\bm{x}^{(t+1)}$ of the next layer. In the paper, we set $V_{\rm th}$ as 1.

There is a notorious problem in SNN training the firing function is undifferentiable.
To demonstrate this problem, we formulate the gradient by the chain rule, given as
\begin{equation}\label{eq:gradiet}
\begin{split}
 \frac{\partial {L}}{\partial {\mathbf{W}}} =  \sum_t (\frac{\partial {L}}{\partial \bm{o}^{(t)}} \frac{\partial {\bm{o}^{(t)}}}{\partial {\bm{u}^{(t), \text{pre}}}} +  \frac{\partial {L}}{\partial {\bm{u}^{(t+1), \text{pre}}}} \frac{\partial {\bm{u}^{(t+1), \text{pre}}}}{\partial {\bm{u}^{(t), \text{pre}}}} )
 \frac{\partial {\bm{u}^{(t), \text{pre}}}}{\partial {\mathbf{W}}}.
 \end{split}
\end{equation}
Since the firing function~\autoref{eq_fire} is similar to the $\mathrm{sign}$ function. The $\frac{\partial {\bm{o}^{(t)}}}{\partial {\bm{u}^{(t), \text{pre}}}}$ is 0 almost everywhere except for the threshold. 
Therefore, the updates for weights would either be 0 or infinity if we use the actual gradient of the firing function.

\section{Methodology}

\subsection{The Gradient Vanishing Problem for SNNs} 
\label{sec_inf}
\begin{figure*} [tp]
	\centering
	\subfloat[\label{fig:a}]{
		\includegraphics[scale=0.12]{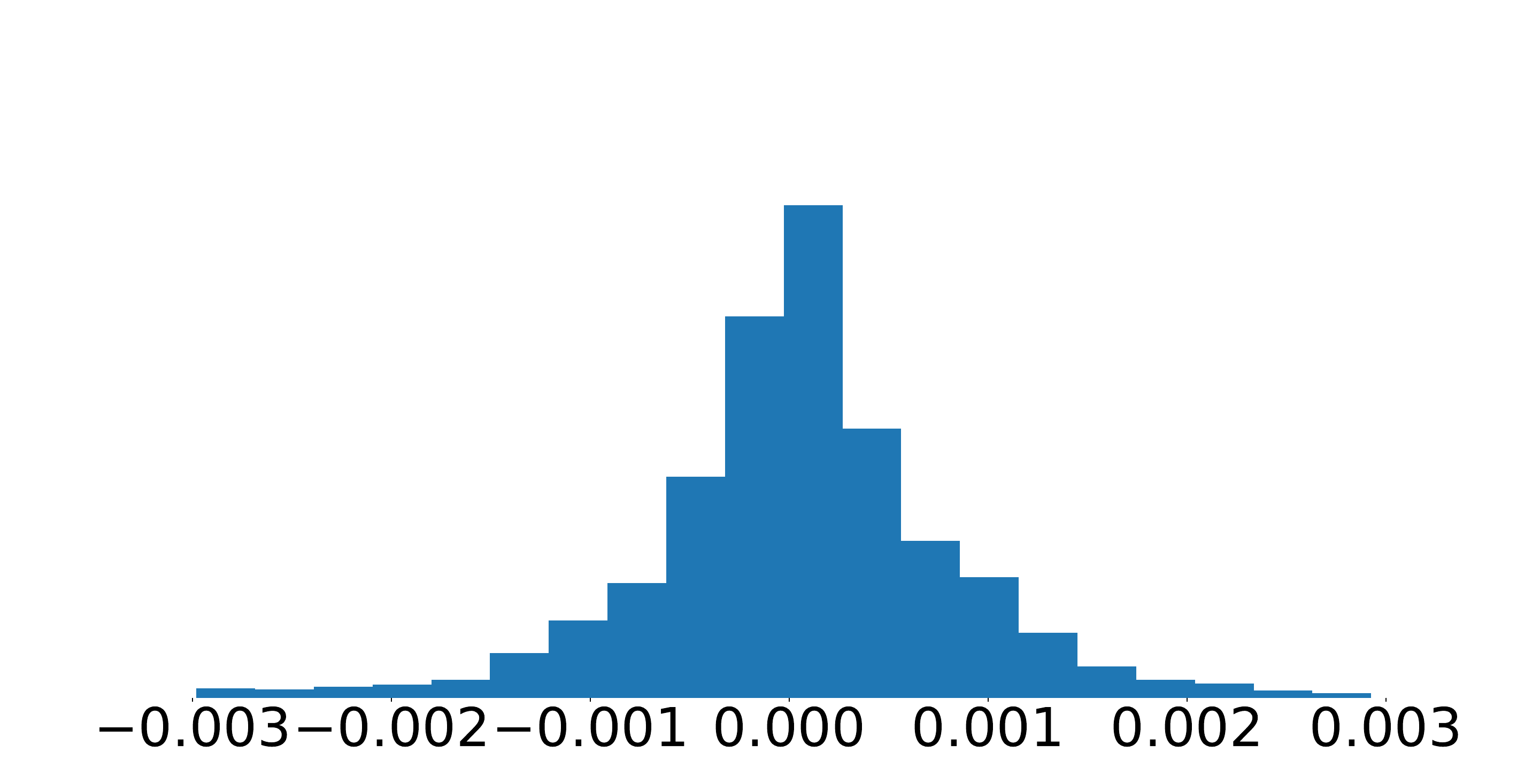}}
	\subfloat[\label{fig:b}]{
		\includegraphics[scale=0.12]{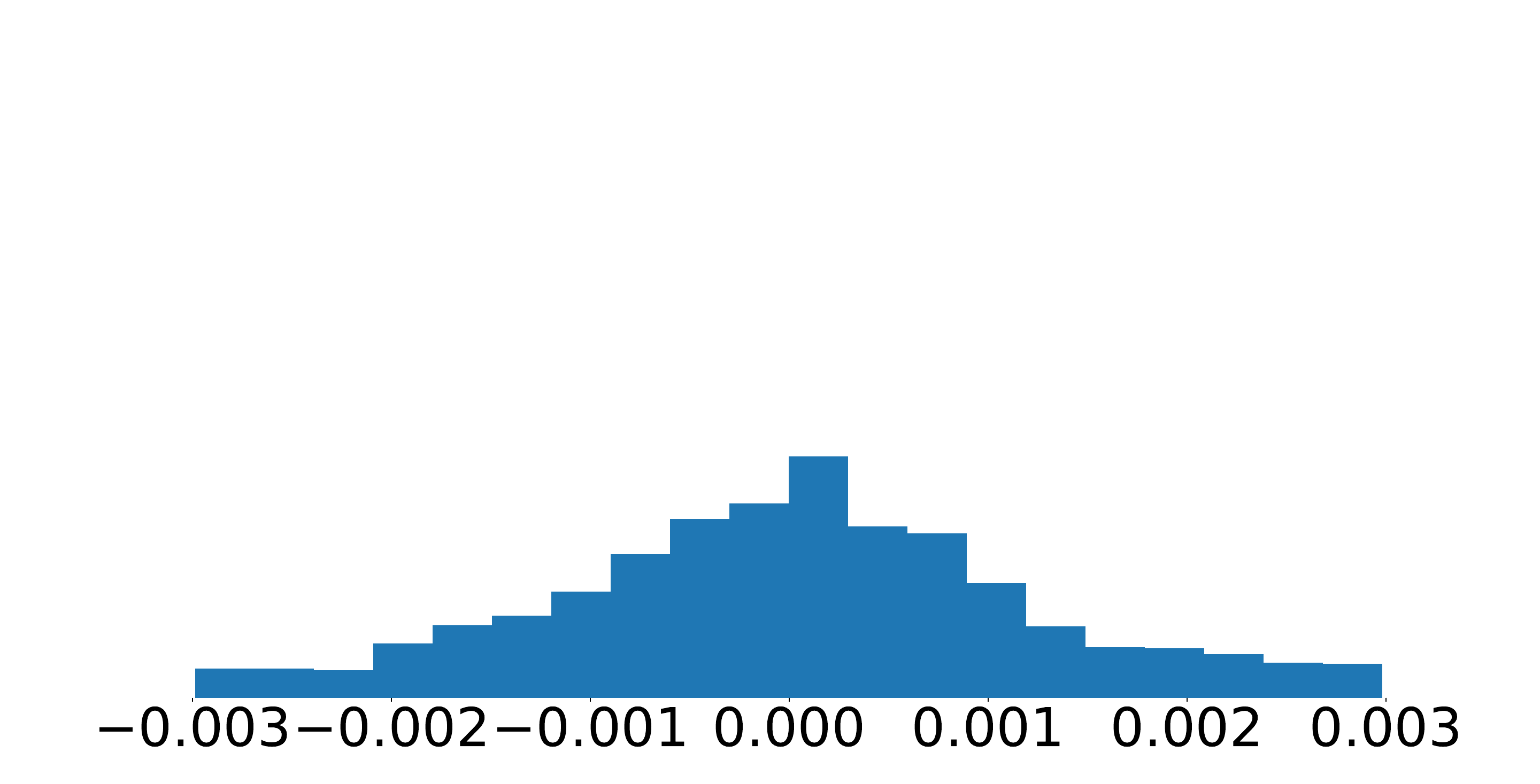}}
	\caption{The gradient distributions of the first layer for Spiking ResNet34 on CIFAR-100. (a) and (b) show the distributions for the vanilla model and the one with the shortcut back-propagation method.}
	\label{gradient} 
\end{figure*}

As aforementioned, the non-differentiability of SNNs poses challenges when training them directly. To address this issue, researchers have proposed the use of surrogate gradients. In this approach, the firing function remains unchanged during the forward pass, but a surrogate function is employed during the backward pass. The surrogate gradient is then computed based on this surrogate function.
There are three commonly used surrogate gradients:
\begin{equation}
\begin{cases}
    \frac{\partial {\bm{o}^{(t)}}}{\partial {\bm{u}^{(t), \text{pre}}}} & =
    \gamma \max\left(0, 1 - \left|\frac{\bm{u}^{(t), \text{pre}}}{V_{\rm th}} - 1\right|\right), \\
    \frac{\partial {\bm{o}^{(t)}}}{\partial {\bm{u}^{(t), \text{pre}}}} & = 
    \frac{1}{a} \mathrm{sign}\left(\left|\bm{u}^{(t), \text{pre}} - V_{\rm th}\right| < \frac{a}{2}\right),\\
    \frac{\partial {\bm{o}^{(t)}}}{\partial {\bm{u}^{(t), \text{pre}}}} & = k (1 - \tanh (\bm{u}^{(t), \text{pre}}- V_{\rm th}))^2.
\end{cases}
\end{equation}
Each of these surrogate gradients includes a hyperparameter that controls the sharpness and width of the gradient. However, it is evident that these gradients, or their approximations, often become close to zero over a significant portion of their intervals. Consequently, this poses a considerable challenge in terms of severe gradient vanishing.
While residual blocks have proven effective in mitigating gradient vanishing problems in traditional neural networks, their performance is not optimal when applied to SNNs. To demonstrate this, we express the skip connection using the following formulation:
\begin{equation}
    \bm{o} = g(f(\bm{x}) + \bm{x}),
\end{equation}
where $f(\cdot)$ is convolutional layers and $g(\cdot)$ is the activation function. The standard ResNet network is composed of multiple skip connection blocks cascaded together. In ANN, ReLU is used for $g(\cdot)$, since ReLU is unbounded for the positive part, the gradient can be passed to the input of the block directly. However, in the case of LIF neurons in SNNs, the gradient will be reduced through the surrogate gradient. Thus, the standard skip connections still suffer the gradient vanishing problem in SNNs.
To visually illustrate this problem, we show the gradient distributions of the first layer for Spiking ResNet34 with 4 timesteps on the CIFAR-100 in the~\autoref{gradient}(a). It can be seen that these gradients are close to 0 in most intervals, meaning the gradient vanishing problem is very significant for these shallow layers. 

\subsection{The Shortcut Back-propagation Method}
\label{shortcut }

Both theoretical analysis and experiments reveal the severity of the gradient vanishing problem in shallow layers of SNNs. In this paper, we propose a shortcut back-propagation method to address this issue. Specifically, the network is divided into several blocks, and we add multiple shortcut branches directly from these blocks to the output, as shown in \autoref{workflow}. These blocks are then followed by a global average pooling layer and a fully connected layer, resulting in the final output:
\begin{equation}
    \bm{o}_{\rm final} = \sum_l b_l(\bm{x}),
\end{equation}
where the $b_l(\bm{x})$ represents the output of the $l$-th block. While the original final output can be expressed as
\begin{equation}
    \bm{o}_{\rm final} = b_n(\bm{x}),  \ \ \text{where } b_l(\bm{x})=f_l(b_{l-1}(\bm{x})).
\end{equation}
In the above equation, the $n$ is the total number of the blocks and $f_l(\cdot)$ is the network of the $l$-th block.
To demonstrate how our method alleviates the gradient vanishing problem, let's consider the gradient of the weight in the first layer as an example. For the original case, it can be expressed as
\begin{equation}
 \frac{\partial {L}}{\partial {\mathbf{W}_1}} =  \frac{\partial {L}}{\partial b_n(\bm{x})} \frac{\partial b_n(\bm{x})}{\partial b_{n-1}(\bm{x})} \cdots
 \frac{\partial b_{l+1}(\bm{x})}{\partial b_{l}(\bm{x})} 
 \frac{\partial b_l(\bm{x})}{\partial {\mathbf{W}_1}}.
\end{equation}
Since the $\frac{\partial b_{l+1}(\bm{x})}{\partial b_{l}(\bm{x})}$ is always reduced through the surrogate gradient, the $\frac{\partial {L}}{\partial {\mathbf{W}_1}} $ becomes very small, resulting in insufficient weight updates.
However, for our method, it can be expressed as
\begin{equation}\label{eq:finalw}
 \frac{\partial {L}}{\partial {\mathbf{W}_1}} =  \sum_l \frac{\partial {L}}{\partial b_l(\bm{x})} 
 \frac{\partial b_l(\bm{x})}{\partial {\mathbf{W}_1}}.
\end{equation}
In our method, the gradient is directly fed into the first block and subsequently to $\mathbf{W}_1$. This completely solves the gradient vanishing problem. To further illustrate this advantage, we visualize the gradient distribution of the first layer for Spiking ResNet34 on CIFAR-100 in \autoref{gradient}(b). It can be observed that the distribution is relatively flat, indicating that the gradient vanishing problem has been effectively addressed in these shallow layers. Moreover, these shortcut branches can be removed during inference, incurring no additional cost.
\begin{algorithm}[t]
	\caption{Training and inference procedure of SNN with our method.}
	\label{alg:algorithm}
    \textcolor{blue}{\textbf{Training}}\\
	\textbf{Input}: An SNN to be trained; Initial balance coefficient $\lambda$; training dataset; total training iteration: $I_{\rm train}$.\\
	\textbf{Output}: The well-trained SNN.
	\begin{algorithmic}[1] 
		\FOR {all $i = 1, 2, \dots , I_{\rm train}$ iteration}
		\STATE Get mini-batch training data, $\bm{x}_{\rm in}(i)$ and class label, $\bm{y}(i)$;
		\STATE Feed the  $\bm{x}_{\rm in}(i)$ into the SNN and calculate every block output $b_l(\bm{x}_{\rm in}(i))$ and the final main net output $b_n(\bm{x}_{\rm in}(i))$;
    	\STATE Update $\lambda$;
  	\STATE Calculate the final output by \autoref {eq:final};
		\STATE Compute classification loss $L_{\rm CE}={\mathcal{L}_{\rm CE}}(\bm{o}_{\rm final}(i),\bm{y}(i))$;
		\STATE Calculate the gradient w.r.t. $\mathbf{W}$ by \autoref{eq:finalw};
		\STATE Update $\mathbf{W}$: $(\mathbf{W} \leftarrow  \mathbf{W} - \eta \frac{\partial {L}}{\partial {\mathbf{W}}})$ where $\eta$ is learning rate.
		\ENDFOR\\
	\end{algorithmic}
    \textcolor{blue}{\textbf{Inference}}\\
	\textbf{Input}: The  trained SNN; test dataset; total test iteration: $I_{\rm test}$.\\
	\textbf{Output}: The result.
	\begin{algorithmic}[1] 
		\FOR {all $i = 1, 2, \dots , I_{\rm test}$ iteration}
		\STATE Get mini-batch test data, $\bm{x}_{\rm in}(i)$ and class label, $\bm{y}(i)$ in test dataset;
        \STATE Feed the  $\bm{x}_{\rm in}(i)$ into the trained SNN;
        \STATE Calculate the final main net output $b_n(\bm{x}_{\rm in}(i))$;
		\STATE Calculate the final output, $\bm{o}_{\rm final}(i) = b_n(\bm{x}_{\rm in}(i))$  ;
		\STATE   Compare the final output $\bm{o}_{\rm final}(i)$ and $\bm{y}(i)$ to compute the classification result.
		\ENDFOR\\
	\end{algorithmic}
\end{algorithm}

\subsection{The Evolutionary Training Framework}

While the shortcut back-propagation method effectively addresses the gradient vanishing problem, it introduces a potential conflict. Each shortcut branch contributes to the final output, but if we focus too much on the shallow layer outputs, the overall accuracy may suffer. This is because the shortcut branches are removed during the inference phase, and the final accuracy is primarily influenced by the main branch. On the other hand, if we prioritize the main branch output, the final loss may not capture enough information from the shallow layers. Consequently, the weights of these shallow layers may not be updated adequately.

\begin{table}[tp]
	\centering	
	\caption{Ablation study for the shortcut back-propagation method.}	
	\label{tab:ab}
  \resizebox{0.65\textwidth}{!}{
	\begin{tabular}{llcc}	
		\toprule
		 Architecture & Method & Time-step & Accuracy \\	
		\toprule
		 \multirow{6}{*}{ResNet18} & Vanilla Training & 2 & 71.42\%   \\	
		                          &  Shortcut Back-propagation & 2 & \textbf{73.68\%}   \\
		                        &  Evolutionary Training & 2 & \textbf{74.02\%}   \\                            
		\cline{2-4}
		 &  Vanilla Training & 4 & 72.22\%   \\	
		                          &  Shortcut Back-propagation & 4 & \textbf{74.78\%}   \\  
		                        &  Evolutionary Training & 4 & \textbf{74.83\%}   \\                              
		\cline{1-4}
		  \multirow{6}{*}{ResNet34}&  Vanilla Training & 2 & 69.82\%   \\	
		                            &  Shortcut Back-propagation & 2 & \textbf{74.06\%}   \\
		                            &  Evolutionary Training & 2 & \textbf{74.17\%}   \\                            
		\cline{2-4}
		   &  Vanilla Training & 4 & 69.98\%   \\	
		                             &  Shortcut Back-propagation & 4 & \textbf{75.67\%}   \\  
		                           &  Evolutionary Training & 4 & \textbf{75.81\%}   \\     	
		\bottomrule			   		         	            			         	
	\end{tabular}}
\end{table}

To address this issue, we propose an evolutionary training framework. During the early stages of training, we prioritize the former side branch net, allowing for sufficient weight updates in the shallow layers. As training progresses, we gradually shift our focus to the main branch net until all attention is on the main net at the end of training. 
To achieve this, we introduce a balance coefficient, denoted by 
 $\lambda (i)$ and adopt a strategy
of decreasing to adjust it as follows,
\begin{equation}\label{eq:final}
    \bm{o}_{\rm final} = b_n(\bm{x}) + \lambda (i)\sum_{l=1} b_l(\bm{x}),  \ \ \text{where } \lambda (i) = \lambda (1 - \frac{i}{I}).
\end{equation}
In the equation mentioned above, $I$ represents the total number of training iterations, $i$ denotes the current training iteration, and $\lambda$ is a constant. In our work, we set $\lambda$ to a value of 0.25.
The training and inference of our SNN  are detailed in
Algo.~\ref{alg:algorithm}.

\section{Experiment}

We conduct extensive experiments on CIFAR-10(100)~\cite{CIFAR-10}, ImageNet~\cite{2009ImageNet}, and CIFAR10-DVS~\cite{2017CIFAR10} to demonstrate the superior performance of our method. The CIFAR-10(100) dataset comprises 50k training images and 10k test images, divided into 10(100) classes, each with $32\times 32$ pixels. CIFAR10-DVS is a converted dataset derived from CIFAR-10. It consists of 10k images, with 1k images per class, in 10 classes. ImageNet is a significantly larger dataset, containing over 1,250k training images and 50k test images. For these static datasets (CIFAR-10, CIFAR-100, and ImageNet), we applied data normalization to ensure that they have $0$ mean and $1$ variance. Additionally, to prevent overfitting, we performed random horizontal flipping and cropping on all these datasets. For a fair comparison, AutoAugment~\cite{cubuk2019autoaugment} was also used for data augmentation following these work~\cite{Guo2022eccv,li2021differentiable} on CIFAR-10(100).  
Regarding the CIFAR10-DVS dataset, we partitioned it into 9k training images and 1k test images, as described in~\cite{2018Direct}. The training image frames were resized to $48\times 48$ as in~\cite{2020Going}. Random horizontal flipping and random roll within a range of 5 pixels were also applied during training. For the test images, we simply resized them to 
 $48\times 48$ without any additional processing, following the approach of Li et al.~\cite{li2021differentiable}.
We run the model with 8 V100.


\begin{table*}[!h]
 \begin{center}
	\caption{Comparison with SoTA methods on CIFAR-10(100).}	
	\label{tab:Comparisoncifar}	
  \resizebox{0.95\textwidth}{!}{
	\begin{tabular}{cllccc}	
		\toprule
		Dataset & Method & Type & Architecture & Timestep & Accuracy \\	
		\toprule
		\multirow{26}{*}{\rotatebox{90}{CIFAR-10}}	
		& TL~\cite{wu2021tandem} & Tandem Learning & CIFARNet & 8 & 89.04\%   \\   
  		&PTL~\cite{wu2021progressive} & Tandem Learning & VGG11 & 16 &  91.24\%   \\ 
		& PLIF~\cite{2020Incorporating} & SNN training & PLIFNet & 8 & 93.50\%   \\
		& DSR~\cite{meng2022training} & SNN training & ResNet18 &  20 & 95.40\%   \\ 
		& KDSNN~\cite{xu2023constructing} & SNN training & ResNet18 &  4 & 93.41\%   \\  
  & \multirow{1}{*}{RecDis-SNN~\cite{Guo_2022_CVPR}} & \multirow{1}{*}{SNN training} & \multirow{1}{*}{ResNet19} 
		& 2 & 93.64\%   \\
		\cline{2-6}
		& \multirow{2}{*}{Diet-SNN~\cite{2020DIET}} & \multirow{2}{*}{SNN training} &  \multirow{2}{*}{ResNet20} & 5 & 91.78\%   \\ 
		&  &  &  & 10 & 92.54\%   \\   
		\cline{2-6}
		& \multirow{2}{*}{Dspike~\cite{li2021differentiable}} & \multirow{2}{*}{SNN training} & \multirow{2}{*}{ResNet20} 
		& 2 & 93.13\%   \\
		&  &  &											                                  & 4 & 93.66\%   \\
		\cline{2-6}
		& \multirow{2}{*}{STBP-tdBN~\cite{2020Going}} & \multirow{2}{*}{SNN training} & \multirow{2}{*}{ResNet19} 
		& 2 & 92.34\%   \\
		&  &  &											                                  & 4 & 92.92\%   \\
		\cline{2-6}
		& \multirow{2}{*}{TET~\cite{deng2022temporal}} & \multirow{2}{*}{SNN training} & \multirow{2}{*}{ResNet19} 
		& 2 & 94.16\%   \\
		&  &  &											                                  & 4 & 94.44\%   \\
		\cline{2-6}
		
		& \multirow{2}{*}{{Real Spike~\cite{guo2022real}}} & \multirow{2}{*}{SNN training} & \multirow{1}{*}{ResNet19} 
		& 2 & {95.31\%} \\
		\cline{4-6}	
		&  &  & \multirow{1}{*}{ResNet20} 		                                          & 4 & {91.89\%}  \\
		\cline{2-6}
		& \multirow{4}{*}{\textbf{Shortcut Back-propagation}} & \multirow{4}{*}{SNN training} & \multirow{2}{*}{ResNet18} 
		& 2 & \textbf{93.89\%}$\pm 0.11$  \\
		&  &  &											                                  & 4 & \textbf{94.30\%}$\pm 0.09$  \\
		\cline{4-6}	
		&  &  & \multirow{2}{*}{ResNet19} 		                                          & 1 & \textbf{94.47\%}$\pm 0.09$   \\
		&  &  &											                                  & 2 & \textbf{95.19\%}$\pm 0.10$   \\
		\cline{2-6}
		& \multirow{4}{*}{\textbf{Evolutionary Training}} & \multirow{4}{*}{SNN training} & \multirow{2}{*}{ResNet18} 
		& 2 & \textbf{93.92\%}$\pm 0.08$  \\
		&  &  &											                                  & 4 & \textbf{94.46\%}$\pm 0.11$  \\
		\cline{4-6}	
		&  &  & \multirow{2}{*}{ResNet19} 		                                          & 1& \textbf{94.81\%}$\pm 0.13$   \\
		&  &  &											                                  & 2 & \textbf{95.36\%}$\pm 0.10$   \\  
\hline
		\multirow{19}{*}{\rotatebox{90}{CIFAR-100}}	
		& T2FSNN~\cite{2020T2FSNN} & ANN2SNN & VGG16 & 680 & 68.80\%   \\ 
		& Real Spike~\cite{guo2022real} & SNN training & ResNet20 & 5 & 66.60\%   \\
  	& LTL~\cite{yang2022training} & Tandem Learning & ResNet20 & 31 & 76.08\%   \\ 
		& \multirow{1}{*}{Diet-SNN~\cite{2020DIET}} & \multirow{1}{*}{SNN training} & ResNet20 & 5 & 64.07\%   \\   
		& \multirow{1}{*}{RecDis-SNN~\cite{Guo_2022_CVPR}} & \multirow{1}{*}{SNN training} & ResNet19 & 4 & 74.10\%   \\   
		\cline{2-6}
		& \multirow{2}{*}{Dspike~\cite{li2021differentiable}} & \multirow{2}{*}{SNN training} & \multirow{2}{*}{ResNet20} 
		& 2 & 71.68\%   \\
		&  &  &											                                  & 4 & 73.35\%   \\
		\cline{2-6}
		& \multirow{2}{*}{TET~\cite{deng2022temporal}} & \multirow{2}{*}{SNN training} & \multirow{2}{*}{ResNet19} 
		& 2 & 72.87\%   \\
		&  &  &											                                  & 4 & 74.47\%   \\
		\cline{2-6}

		& \multirow{4}{*}{\textbf{Shortcut Back-propagation}} & \multirow{4}{*}{SNN training} & \multirow{2}{*}{ResNet18} 
		& 2 & \textbf{73.68\%}$\pm 0.10$  \\
		&  &  &											                                  & 4 & \textbf{74.78\%}$\pm 0.08$  \\
		\cline{4-6}	
		&  &  & \multirow{2}{*}{ResNet19} 		                                          & 1 & \textbf{75.75\%}$\pm 0.10$   \\
		&  &  &											                                  & 2 & \textbf{77.56\%}$\pm 0.13$   \\
		\cline{2-6}
		& \multirow{4}{*}{\textbf{Evolutionary Training}} & \multirow{4}{*}{SNN training} & \multirow{2}{*}{ResNet18} 
		& 2 & \textbf{74.02\%}$\pm 0.09$  \\
		&  &  &											                                  & 4 & \textbf{74.83\%}$\pm 0.11$  \\
		\cline{4-6}	
		&  &  & \multirow{2}{*}{ResNet19} 		                                          & 1 & \textbf{75.82\%}$\pm 0.12$   \\
		&  &  &											                                  & 2 & \textbf{77.79\%}$\pm 0.08$   \\ 
  
		\bottomrule				         	
	\end{tabular}}	
 \end{center}
\end{table*}
\subsection{Ablation Study}
To validate the effectiveness of the proposed shortcut back-propagation method, we initially conducted several ablation experiments on the CIFAR-100 dataset using ResNet18 and ResNet34 with different timesteps. The results are detailed in \autoref{tab:ab}. For ResNet18, the accuracy achieved through vanilla training is 71.42\% and 72.22\% under 2 and 4 timesteps, respectively, which aligns with existing works. Upon applying our shortcut back-propagation method, the accuracy improved to 73.68\% and 74.78\%, marking a notable 2.5\% enhancement. Furthermore, with the evolutionary training method, the performance of ResNet18 saw an additional improvement, reaching 74.02\% and 74.83\%, respectively.
Under vanilla training, ResNet34 achieved accuracies of 69.82\% and 69.98\% with 2 and 4 timesteps, respectively. These results are actually worse than those obtained with ResNet18. This suggests that the deeper model does not exhibit better performance due to the significant gradient vanishing problem in SNNs.
However, by utilizing our shortcut back-propagation method, the accuracy significantly improves to 74.06\% and 75.67\%, representing a remarkable 5.0\% enhancement. Notably, these results surpass the performance of ResNet18 as well. This clearly demonstrates the effectiveness of our proposed method.
Furthermore, when incorporating the evolutionary training method, we observe further improvements in performance. 

\begin{table*}[t]
	\centering	
	\caption{Comparison with SoTA methods on ImageNet.}	
	\label{tab:Comparisonimage}	
 \resizebox{0.8\textwidth}{!}{
	\begin{tabular}{llccc}	
		\toprule
		Method & Type & Architecture & Timestep &  Accuracy \\	
		\toprule
				
		TET~\cite{deng2022temporal} &  SNN training & ResNet34 & 6 & 64.79\%   \\
		RecDis-SNN~\cite{Guo_2022_CVPR} &  SNN training & ResNet34 & 6 & 67.33\%   \\
        GLIF~\cite{yao2023glif} & SNN training & ResNet34 &  4 & 67.52\%   \\ 
        DSR~\cite{meng2022training} & SNN training & ResNet18 &  50 & 67.74\%   \\
        MS-ResNet~\cite{hu2023advancing} & SNN training & ResNet18 &  6 & 63.10\%   \\
        \cline{1-5}
		\multirow{2}{*}{MPBN~\cite{Guo_2023_ICCV}} & \multirow{2}{*}{SNN training} & {ResNet18} & 4  & {63.14\%}   \\
		 &  &											                             {ResNet34} & 4  & {64.71\%}   \\ 
		\cline{1-5}
		\multirow{2}{*}{Real Spike~\cite{guo2022real}} & \multirow{2}{*}{SNN training} & {ResNet18} & 4  & {63.68\%}   \\
		 &  &											                             {ResNet34} & 4  & {67.69\%}   \\ 
		\cline{1-5}		
		\multirow{2}{*}{SEW ResNet~\cite{2021Deep}} & \multirow{2}{*}{SNN training} & {ResNet18} & 4  & {63.18\%}   \\
		 &  &											                             {ResNet34} & 4  & {67.04\%}   \\ 
		\cline{1-5}
		\multirow{2}{*}{\textbf{Shortcut Back-propagation} } & \multirow{2}{*}{SNN training} & {ResNet18} & 4 & \textbf{64.47\%}$\pm 0.21$   \\
		 &  &											                             {ResNet34} & 4& \textbf{67.90\%}$\pm 0.17$   \\
		\cline{1-5}
		\multirow{2}{*}{\textbf{Evolutionary Training} } & \multirow{2}{*}{SNN training} & {ResNet18} & 4 & \textbf{65.12\%}$\pm 0.18$   \\
		 &  &											                             {ResNet34} & 4& \textbf{68.14\%}$\pm 0.15$   \\   
		\bottomrule				         	
	\end{tabular}}
\end{table*}

\subsection{Comparison with SoTA Methods}

In this section, we conducted a comparative experiment for the shortcut back-propagation method and the evolutionary training framework, taking into consideration several state-of-the-art works. To ensure a fair comparison, we present the top-1 accuracy results based on 3 independent trials. 
We first evaluated our method on CIFAR-10 and CIFAR-100 datasets. 
The AdamW optimizer was employed with a learning rate of 0.01 which is cosine decay to 0 and a weight decay of 0.02. Throughout the training process, all models were trained using a batch size of 128 for a total of 300 epochs.
The results are summarized in \autoref{tab:Comparisoncifar}.
For the CIFAR-10 dataset,
we chose SpikeNorm~\cite{2019Going}, Hybrid-Train~\cite{2020Enabling}, TSSL-BP~\cite{2020Temporal}, TL~\cite{wu2021tandem}, PTL~\cite{wu2021progressive}, PLIF~\cite{2020Incorporating}, DSR~\cite{meng2022training}, KDSNN~\cite{xu2023constructing}, Diet-SNN~\cite{2020DIET}, Dspike~\cite{li2021differentiable}, STBP-tdBN~\cite{2020Going}, TET~\cite{deng2022temporal}, RecDis-SNN~\cite{Guo_2022_CVPR}, and Real Spike~\cite{guo2022real} 
as our comparison.
Previous works utilizing ResNet18, ResNet19, and ResNet20 as backbones achieved the highest accuracies of 95.40\%, 95.31\%, and 93.66\% with 20, 2, and 4 timesteps respectively. While our method based on ResNet18 and ResNet19 could reach 93.92\% and 95.36\% with 4 and 2 timesteps, respectively. Note that, ResNet18 is smaller than ResNet20.
On the CIFAR-100 dataset, our method can also achieve better accuracy than other prior state-of-the-art works with fewer timesteps. For instance, our ResNet19 model with only 2 timesteps outperforms the current best method, TET and RecDis-SNN even with 4 timesteps by about 3.3\%. These experimental results clearly demonstrate the efficiency and effectiveness of our method.

\begin{table*}[pt]
	\centering	
	\caption{Comparison with SoTA methods on CIFAR10-DVS.}	
	\label{tab:Comparisondvs}	
\resizebox{0.8\textwidth}{!}{	
 \begin{tabular}{llccc}	
		\toprule
		 Method & Type & Architecture & Timestep & Accuracy \\	
		\toprule
		STBP-tdBN~\cite{2020Going} & SNN training & ResNet19 & 10 & 67.80\%   \\ 
		RecDis-SNN~\cite{Guo_2022_CVPR} & SNN training & ResNet19 & 10 & 72.42\%   \\ 
		 \multirow{1}{*}{Real Spike~\cite{guo2022real}} & \multirow{1}{*}{SNN training} & {ResNet19} 
		& 10 & {72.85\%}   \\
    	Dspike~\cite{li2021differentiable} & SNN training & ResNet18 & 10 & 75.40\%   \\ 
		\cline{1-5}
		\textbf{Shortcut Back-propagation} & SNN training & ResNet18 & 10 & \textbf{82.00\%}$\pm 0.10$   \\ 
		\textbf{Evolutionary Training} & SNN training & ResNet18 & 10 & \textbf{83.30\%}$\pm 0.10$   \\      
		\bottomrule				         	
	\end{tabular}}	
\end{table*}

We proceeded to conduct experiments on the ImageNet dataset, which is a more complex dataset than CIFAR. The learning rate is adjusted to $ 4e^{-3}$ here. The comparative results are presented in \autoref{tab:Comparisonimage}. Notably, there have been several state-of-the-art (SoTA) baselines proposed for this dataset recently, such as RecDis-SNN~\cite{Guo_2022_CVPR}, GLIF~\cite{yao2023glif}, DSR~\cite{meng2022training}, MPBN~\cite{Guo_2023_ICCV}, MS-ResNet~\cite{hu2023advancing}, Real Spike~\cite{guo2022real}, and SEW ResNet~\cite{2021Deep}.
It is important to note that Real Spike and SEW ResNet deviate from the typical ResNet backbone as they generate integer outputs in the intermediate layers, making them more energy-intensive compared to methods with standard backbones. In contrast, our approach adopts the standard ResNet18 and ResNet34 architectures, yet it still outperforms Real Spike and SEW ResNet. Specifically, our method achieves an accuracy of 65.12\% and 68.14\% using ResNet18 and ResNet34, respectively, surpassing Real Spike by 1.44\% and 0.45\%, respectively. This improvement is noteworthy and demonstrates the effectiveness of our method in handling large-scale datasets.

In addition to the aforementioned experiments, we conducted tests on the highly popular neuromorphic dataset, CIFAR10-DVS. We use the same hyper-parameter setting as CIFAR. Employing ResNet18 as the foundational architecture, which is notably smaller compared to ResNet19, our approach achieved remarkable accuracies of 82.00\% and 83.30\%, respectively. These results demonstrate a substantial improvement over previous methodologies.

\section{Conclusion}
In the paper, we proved that the Spiking Neural Network suffers severe gradient vanishing with theoretical justifications and in-depth experimental analysis.
To mitigate the problem, we proposed a shortcut back-propagation method. This enables us to present the gradient to the shallow layers directly, thereby significantly mitigating the gradient vanishing problem.
Additionally, this method does not introduce any burden during the inference phase.
we also presented an evolutionary training framework by inducing a balance coefficient that dynamically changes with the training epoch, which could further improve the accuracy.
We conducted various experiments to verify the effectiveness of our method.

\bibliography{example_paper}
\bibliographystyle{icml2024}

\newpage
\section*{NeurIPS Paper Checklist}

\begin{enumerate}

\item {\bf Claims}
    \item[] Question: Do the main claims made in the abstract and introduction accurately reflect the paper's contributions and scope?
    \item[] Answer:  \answerYes{}.
    \item[] Justification: {We clearly state the claims made and the contributions made in both the abstract and introduction.}

\item {\bf Limitations}
    \item[] Question: Does the paper discuss the limitations of the work performed by the authors?
    \item[] Answer: \answerNA{}.
    \item[] Justification: We find no limitation which we feel must be specifically highlighted here.

\item {\bf Theory Assumptions and Proofs}
    \item[] Question: For each theoretical result, does the paper provide the full set of assumptions and a complete (and correct) proof?
    \item[] Answer: \answerYes{}.
    \item[] Justification: We provide the full set of assumptions and
complete proofs in the Section 4.

    \item {\bf Experimental Result Reproducibility}
    \item[] Question: Does the paper fully disclose all the information needed to reproduce the main experimental results of the paper to the extent that it affects the main claims and/or conclusions of the paper (regardless of whether the code and data are provided or not)?
    \item[] Answer: \answerYes{}.
    \item[] Justification: We provide the detail experiment settings in the Section 5.

\item {\bf Open access to data and code}
    \item[] Question: Does the paper provide open access to the data and code, with sufficient instructions to faithfully reproduce the main experimental results, as described in supplemental material?
    \item[] Answer: \answerYes{}.
    \item[] Justification: We provide open access to the data and code with sufficient instructions in the supplemental material.

\item {\bf Experimental Setting/Details}
    \item[] Question: Does the paper specify all the training and test details (e.g., data splits, hyperparameters, how they were chosen, type of optimizer, etc.) necessary to understand the results?
    \item[] Answer: \answerYes{}.
    \item[] Justification: All implementations are described in the experiments section.

\item {\bf Experiment Statistical Significance}
    \item[] Question: Does the paper report error bars suitably and correctly defined or other appropriate information about the statistical significance of the experiments?
    \item[] Answer: \answerYes{}.
    \item[] Justification: We report the mean as well as the standard deviation accuracy in experiments.

\item {\bf Experiments Compute Resources}
    \item[] Question: For each experiment, does the paper provide sufficient information on the computer resources (type of compute workers, memory, time of execution) needed to reproduce the experiments?
    \item[] Answer: \answerYes{}.
    \item[] Justification: The computation resources
description is provided in the experiment section.
    
\item {\bf Code Of Ethics}
    \item[] Question: Does the research conducted in the paper conform, in every respect, with the NeurIPS Code of Ethics \url{https://neurips.cc/public/EthicsGuidelines}?
    \item[] Answer: \answerYes{}.
    \item[] Justification: The research conducted with the NeurIPS Code of Ethics

\item {\bf Broader Impacts}
    \item[] Question: Does the paper discuss both potential positive societal impacts and negative societal impacts of the work performed?
    \item[] Answer: \answerNo{}.
    \item[] Justification: There is no societal impact of the work performed.
    
\item {\bf Safeguards}
    \item[] Question: Does the paper describe safeguards that have been put in place for responsible release of data or models that have a high risk for misuse (e.g., pretrained language models, image generators, or scraped datasets)?
    \item[] Answer: \answerNA{}.
    \item[] Justification: The paper poses no such risks.

\item {\bf Licenses for existing assets}
    \item[] Question: Are the creators or original owners of assets (e.g., code, data, models), used in the paper, properly credited and are the license and terms of use explicitly mentioned and properly respected?
    \item[] Answer: \answerYes{}.
    \item[] Justification: The original paper for datasets we used are all cited.

\item {\bf New Assets}
    \item[] Question: Are new assets introduced in the paper well documented and is the documentation provided alongside the assets?
    \item[] Answer: \answerNA{}.
    \item[] Justification: We adopt public datasets.

\item {\bf Crowdsourcing and Research with Human Subjects}
    \item[] Question: For crowdsourcing experiments and research with human subjects, does the paper include the full text of instructions given to participants and screenshots, if applicable, as well as details about compensation (if any)? 
    \item[] Answer: \answerNA{}.
    \item[] Justification: The paper does not involve crowdsourcing nor research with human subjects.

\item {\bf Institutional Review Board (IRB) Approvals or Equivalent for Research with Human Subjects}
    \item[] Question: Does the paper describe potential risks incurred by study participants, whether such risks were disclosed to the subjects, and whether Institutional Review Board (IRB) approvals (or an equivalent approval/review based on the requirements of your country or institution) were obtained?
    \item[] Answer: \answerNA{}.
    \item[] Justification: The paper does not involve crowdsourcing nor research with human subjects.

\end{enumerate}

\end{document}